\documentclass{article}
\usepackage{arxiv}
\usepackage[utf8]{inputenc} 
\usepackage[T1]{fontenc}    
\usepackage{hyperref}       
\usepackage{url}            
\usepackage{booktabs}       
\usepackage{amsfonts}       
\usepackage{nicefrac}       
\usepackage{microtype}      
\usepackage{lipsum}		
\usepackage{listings}
\usepackage{xcolor}
\usepackage{booktabs}
\usepackage{caption}
\usepackage{adjustbox}
\usepackage{makecell}
\usepackage{appendix}
\usepackage[numbers]{natbib}

\title{Towards Autoformalization of LLM-generated Outputs for Requirement Verification}

\author{{Mihir Gupte} \\
	General Motors \\
	\texttt{mihir.gupte@gm.com} \\
	\And
	{Ramesh S.} \\
    General Motors \\
	\texttt{ramesh.s@gm.com} \\
}

\renewcommand{\shorttitle}{Technical Report}

\hypersetup{
pdftitle={Towards Autoformalization of LLM-generated Outputs for Requirement Verification},
pdfsubject={},
pdfauthor={Mihir Gupte},
pdfkeywords={Retrieval-Augmented Generation},
}

\begin{document}
\maketitle
\begin{abstract}
\label{sec:abstract}
Autoformalization, the process of translating informal statements into formal logic, has gained renewed interest with the emergence of powerful Large Language Models (LLMs). While LLMs show promise in generating structured outputs from natural language (NL), such as Gherkin Scenarios from NL feature requirements, there's currently no formal method to verify if these outputs are accurate. This paper takes a preliminary step toward addressing this gap by exploring the use of a simple LLM-based autoformalizer to verify LLM-generated outputs against a small set of natural language requirements. We conducted two distinct experiments. In the first one, the autoformalizer successfully identified that two differently-worded NL requirements were logically equivalent, demonstrating the pipeline's potential for consistency checks. In the second, the autoformalizer was used to identify a logical inconsistency between a given NL requirement and an LLM-generated output, highlighting its utility as a formal verification tool. Our findings, while limited, suggest that autoformalization holds significant potential for ensuring the fidelity and logical consistency of LLM-generated outputs, laying a crucial foundation for future, more extensive studies into this novel application.
\end{abstract}


\section{Introduction}
\label{sec:intro}
\label{sec:introduction}

As Language Model-based (LLM) applications become more prevalent, there's an increased need for a framework to verify their outputs. Concurrently, the field of autoformalization\cite{weng2025autoformalization}, the process of translating informal statements into formal logic, has gained significant traction, fueled by LLMs' impressive ability to translate informal statements into formats like first-order logic.

\begin{figure}[h] 
    \centering
    \includegraphics[width=1\linewidth]{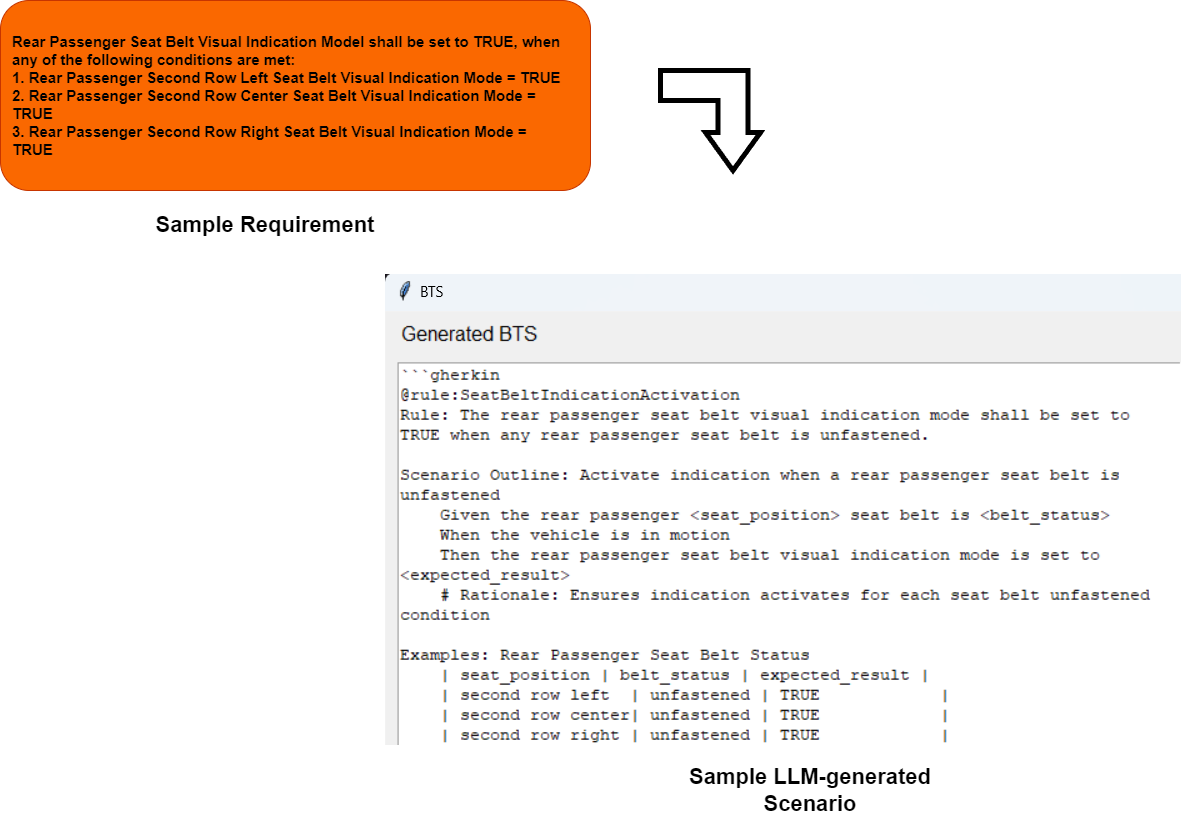} 
    \caption{Sample Output generated by a LLM-powered tool on a Natural Language requirement }
    \label{fig:llm-powered-generation}
\end{figure}

In this paper, we explore the outputs of a sample tool used for building LLM-powered applications to automate the generation of test case scenarios from natural language requirements. For example, a requirement such as: \textit{"Given the vehicle speed $\geq$ 10, When the seatbelt is unfastened, then initiate Seatbelt Reminder Chime"} can be used to generate a structured Gherkin scenario. However, a major problem arises: there is no formal method to verify if the LLM-generated scenario accurately reflects the original requirement. Figure \ref{fig:llm-powered-generation} illustrates an example of an LLM-generated output for a specific requirement. While these tools show impressive results, a critical gap remains: there is no way to formally verify if the generated scenario matches the given requirement.

This is where Autoformalization presents a promising solution\cite{wu2022autoformalization}. By translating both the natural language requirement and the LLM-generated output into a formal logical representation, we can use formal verification tools to check for logical equivalence. For this paper, we explore this potential using DeepSeek-Prover\cite{ren2025deepseek}, an LLM fine-tuned on formal mathematical data, to perform this verification.
\newpage
In this paper, we take the first steps to explore a novel application of autoformalization in a limited scope:
\begin{enumerate}
    \item We demonstrate a basic pipeline using DeepSeek-Prover to verify if an LLM-generated output correctly matches its corresponding natural language requirement.
    \item We explore a pipeline to determine if two differently written natural language requirements are logically equivalent and point to the same scenario.
\end{enumerate}

\section{Related Work}
\label{sec:related_work}
Recent work on autoformalization has largely focused on two key areas: \textbf{creating high-quality datasets} and \textbf{developing new strategies for large language models (LLMs) training} to handle formal logic.

On the data front, researchers have created several specialized datasets to facilitate the training of autoformalizer models. Notable examples include ProofNet \cite{azerbayev2023proofnet} and MiniF2F \cite{zhengminif2f}. Despite these efforts, a major challenge across the field remains the scarcity of high-quality data due to its highly specific and technical nature. While many early datasets focused on a single formal language like Lean \cite{moura2021lean}, recent studies by \cite{jiang2024multi} found that training models on multiple languages, such as Lean, Isabelle \cite{paulson1994isabelle}, and Coq \cite{huet1997coq}, can actually improve performance. This has led to a focus on new methods to increase data quantity and diversity \cite{liu2025atlas}.

In parallel, a significant portion of the research landscape is dedicated to developing more effective strategies for LLMs to tackle formal problems. A key innovation was the "draft, sketch, and prove" method proposed by \cite{jiang2022draft}, which trains models to create proof outlines as a guide for formalizing theorems. This idea inspired methods that teach models to break down large goals into smaller, more manageable sub-goals, a technique used by models like DeepSeek-Prover \cite{ren2025deepseek}. Other research has explored using more efficient search algorithms to navigate large proof spaces \cite{xin2025bfs} and harnessing reinforcement learning to improve success rates \cite{wang2025kimina, xin2024deepseek}.

Another key area of research addresses the non-deterministic nature of LLMs, which often leads to a significant discrepancy between \textit{pass@1} and \textit{pass@k} results for autoformalizer models \cite{li2024autoformalize}. To alleviate this, a breadth of work focuses on establishing frameworks that utilize symbolic equivalence \cite{li2024autoformalize} or employ Retrieval-Augmented Generation (RAG) methods to retrieve precise formal definitions for contextual grounding \cite{lu2025automated}. From a different perspective, autoformalization is also being used as a framework to verify variations of solutions generated by LLMs for mathematical problems \cite{zhou2024don}.

Apart from verifying formal statements from natural language, a significant amount of work focuses on using LLMs to solve mathematical problems at all levels, from high school \cite{zhengminif2f} to undergraduate-level mathematics \cite{azerbayev2023proofnet}. Mathematics is a primary focus for autoformalization as the problem extends naturally to formal mathematical proofs \cite{yang2024formal}. This research also branches into other domains, including geometry-based problems \cite{murphy2024autoformalizing} and partial differential equations for physics \cite{soroco2025pde}. However, the potential for autoformalization extends beyond traditional academic fields, with significant scope in areas such as code-based verification \cite{cunningham2022towards} and even biomedical fact verification \cite{hamed2025knowledge}.

\section{Background}
\label{sec:background}
\subsection{LLM for Formal Theorem Proving}
As high-quality datasets for formal mathematical theorems have become more widely adopted, researchers have begun fine-tuning language models for the sole purpose of theorem proving. While LLMs have shown success in generalizing natural language, generalizing formal mathematics remains a significant challenge. However, recent advancements in reasoning-based models have led to considerable improvement in this area.

A key development is the approach proposed by \cite{jiang2022draft}, which uses LLMs to draft proof sketches and decompose complex problems into smaller, more manageable subgoals. This method has become a popular training strategy for autoformalizer models. DeepSeek-Prover-v2 \cite{ren2025deepseek} takes this approach by training the model to perform a chain-of-thought process, recursively searching for proofs via subgoal decomposition and resolution. The model is trained on data from formal languages like Lean, which is crucial for our purpose of converting Natural Language into Lean syntax.

DeepSeek-Prover-v2 comes in two versions: a 7B and a 671B model. For this paper, we use the 7B model due to resource constraints. The model's effectiveness for our task stems from two key factors: its training on the Lean 4 formal language, which is necessary for our pipeline, and its extensive pre-training in natural language, which enables it to effectively translate and formalize requirements.

Other notable models trained for formal mathematical verification include those from \cite{lewkowycz2022solving}, \cite{wang2025kimina} and \cite{azerbayev2023llemma}.

\subsection{Formalizing Theorems of NL Requirements Using Lean}
Lean 4 \cite{moura2021lean} is a powerful theorem prover and formal language. It allows for the expression of complex mathematical and logical statements in a precise, code-like syntax. This makes it the ideal language for most LLM-based formal theorem provers, including DeepSeek-Prover-v2, which is trained to generate proofs in Lean 4. In our experiment, we leverage this capability for two main purposes: to translate natural language (NL) requirements into formal Lean propositions and to formally verify their logical equivalence using theorems.

We use an autoformalizer to convert a given NL requirement into a Lean proposition. Each requirement can be expressed as a logical statement. Given the nature of our requirements (e.g., "If A then B"), they naturally translate into a propositional implication (A → B). The autoformalizer's output is a theorem proposition that can be evaluated as either true or false.

As seen in Figure 2, a given natural language requirement is transformed into a formal Lean statement by our autoformalizer. This formal representation is the foundation for the verification pipelines we describe in the following sections.
\clearpage
\begin{figure}[h] 
    \centering
    \includegraphics[width=0.75\linewidth]{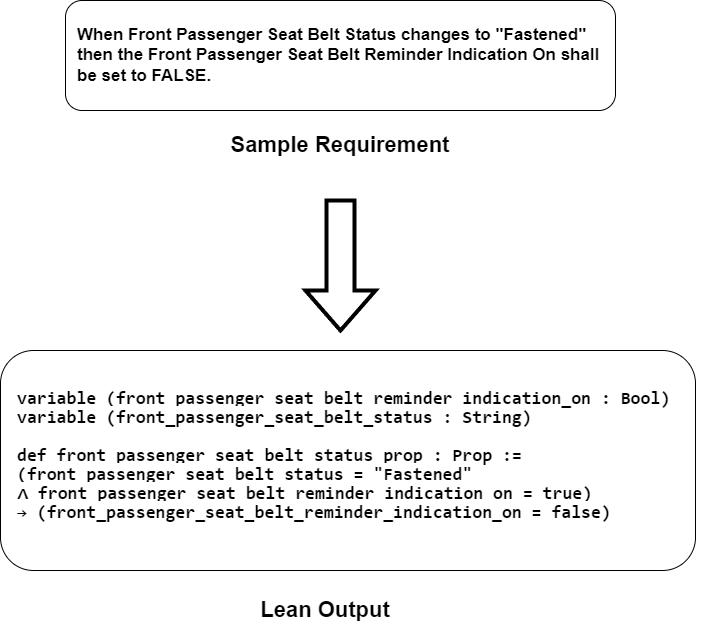} 
    \caption{Sample Lean Output from DeepSeek-Prover-V2 }
    \label{fig:sample-lean}
\end{figure}

\subsection{Biconditional Equivalence of Propositions}
In formal logic, a biconditional or material equivalence is a logical connective that links two propositions and is true only when both propositions share the same truth value. It is represented by the symbol $\leftrightarrow$, and a statement of the form $A \leftrightarrow B$ can be read as "A if and only if B," or simply, "A is equivalent to B".

In our experiment, we use this principle to formally verify the logical equivalence of two statements. We take a pair of natural language statements—either one requirement and one LLM-generated output, or two distinct requirements—and translate each into a formal proposition using our autoformalizer. Let's call these propositions $P_A$ and $P_B$.

The core of our verification process is a theorem-proving task. We instruct the LLM-based theorem prover to prove the biconditional statement $P_A \leftrightarrow P_B$. The model attempts to prove that the propositions are logically equivalent by demonstrating both the forward implication ($P_A \rightarrow P_B$) and the reverse implication ($P_B \rightarrow P_A$). If the model successfully proves the theorem, we can conclude that the original natural language statements are logically consistent. Conversely, if it cannot, it indicates a logical discrepancy between the two.

\section{Methodology}
\label{sec:methods}

We follow a simple workflow for both the experiments that we conduct:
\begin{enumerate}
    \item Formalize given statement pairs in a Propositional Form in Lean using the LLM.
    \item Ground the generated variables in the propositional definitions.
    \item Use the LLM to formally prove the biconditional equivalence of the propositions, then check for logical consistency in Lean.
\end{enumerate}

With this, we elaborate on our methodology in detail in the below sections.

\begin{figure}[h] 
    \centering
    \includegraphics[width=0.75\linewidth]{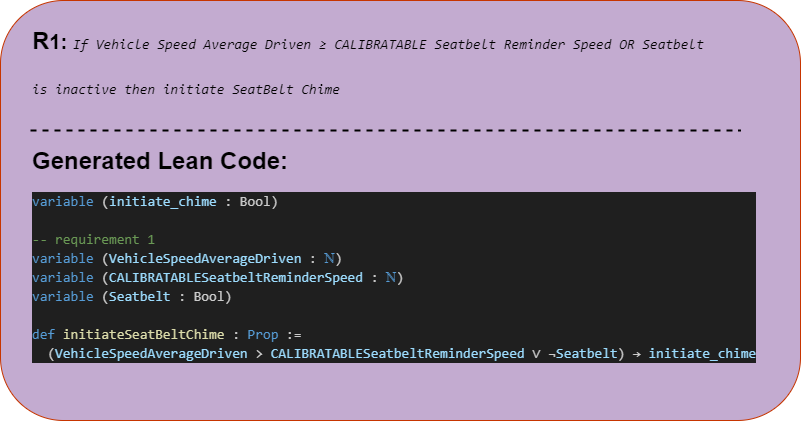} 
    \caption{Lean Formalization of Requirement $R_1$ }
    \label{fig:sample R1 output}
\end{figure}

\begin{figure}[h] 
    \centering
    \includegraphics[width=0.75\linewidth]{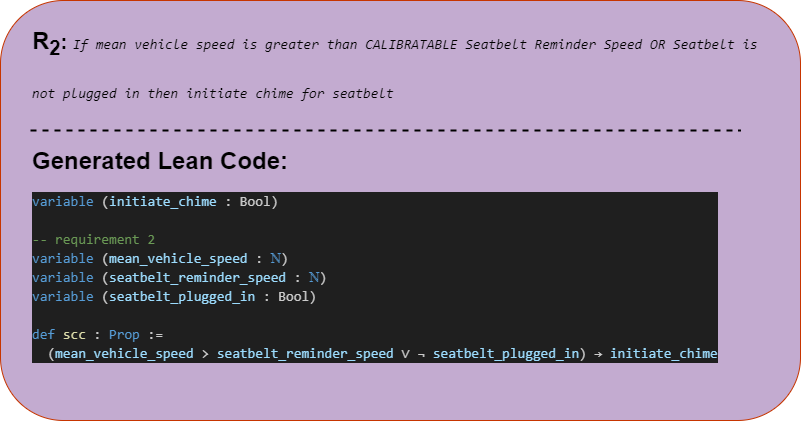} 
    \caption{Lean Formalization of Requirement $R_2$ }
    \label{fig:sample R2 output}
\end{figure}

\subsection{Verifying Logical Equivalence of Requirement Pairs}
Given a set of two natural language requirements:

\textit{$R_1$: If Vehicle Speed Average Driven $\geq$ CALIBRATABLE Seatbelt Reminder Speed OR Seatbelt is inactive then initiate SeatBelt Chime}

\textit{$R_2$: If mean vehicle speed is greater than CALIBRATABLE Seatbelt Reminder Speed OR Seatbelt is not plugged in then initiate chime for seatbelt}

These are two different Natural Language statements pointing to the same requirement, however that is very hard to determine unless extensively checked. To automatically verify if they are logically equivalent, we use our workflow. We can see the output generated by DeepSeek-Prover on $R_1$ and $R_2$ and in Fig \ref{fig:sample R1 output} and Fig \ref{fig:sample R2 output} respectively.

\begin{figure}[h] 
    \centering
    \includegraphics[width=1\linewidth]{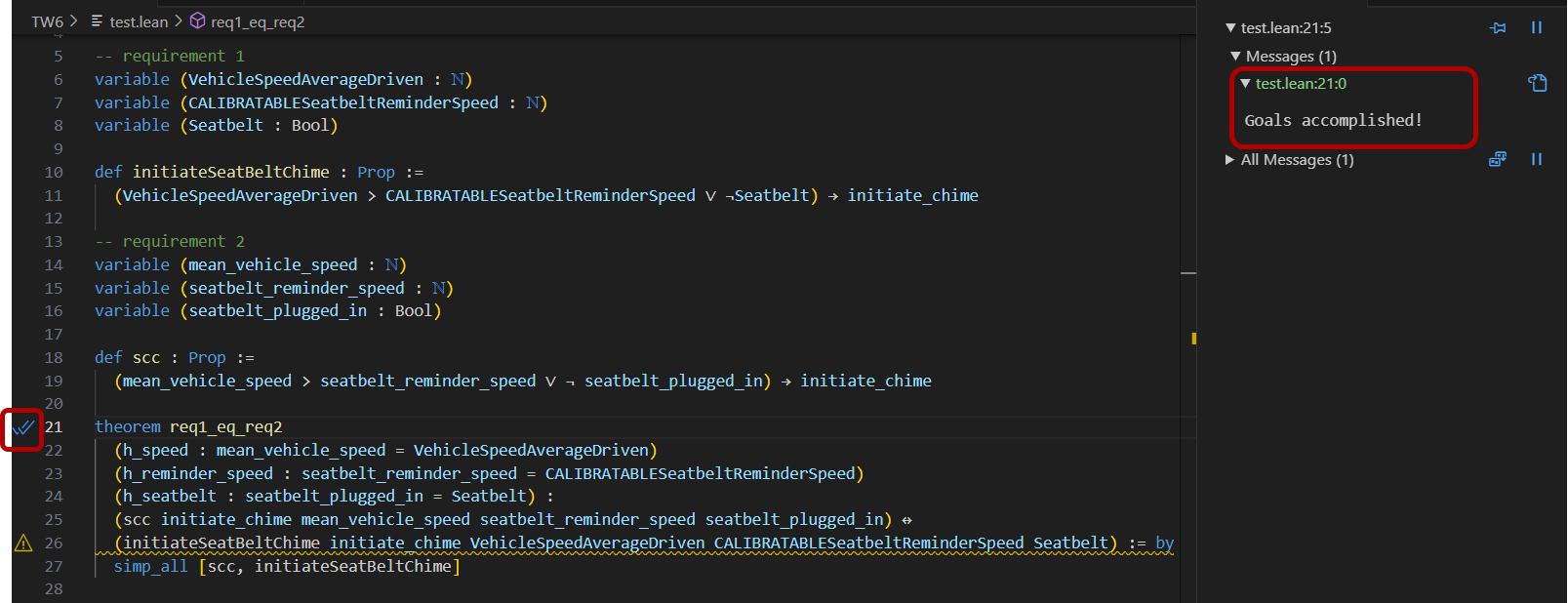} 
    \caption{Using Lean 4 to prove $R_1$ and $R_2$ are logically equivalent }
    \label{fig:r1-r2 logical equivalence}
\end{figure}

We then need to ground the variables. This effort is done manually. By grounding, we need to manually point to the Lean compiler that a set of variables having different naming conventions actually refer to the same variable. There is scope to automate this process using the LLM as well, however due to certain constraints, we choose to do this step manually. Once we specify that, we provide the model with all the information and ask it to either prove the theorem, if possible. The template format we specify to achieve that is provided in Appendix \ref{app:prompt}. The final result of the LLM-generated Lean 4 code in the Lean compiler is provided in Fig \ref{fig:r1-r2 logical equivalence}.

\subsection{Verifying LLM-generated Outputs with NL Requirement}
Given a Natural Language Requirement $R_3$:

\textit{When Front Passenger Seat Belt Status changes to "Fastened" then the Front Passenger Seat Belt Reminder Indication On shall be set to FALSE.}

and a sample LLM-generated Gherkin Output $G_3$ for $R_3$ :

\fbox{
\begin{minipage}{\dimexpr\textwidth-2\fboxsep-2\fboxrule\relax}
Scenario Outline: Seatbelt reminder activation and deactivation based on seatbelt status and seat occupancy\\
Given the front passenger seat is \textless seat\_occupancy\textgreater\\
And the front passenger seat belt status is \textless initial\_seatbelt\_status\textgreater\\
When the front passenger seat belt status changes to \textless final\_seatbelt\_status\textgreater\\
Then the Front Passenger Seat Belt Reminder Indication On shall be set to \textless expected\_reminder\_status\textgreater\\
\\
\begin{tabular}{|c|c|c|c|}
    \hline
    \multicolumn{4}{|c|}{Examples: Seatbelt status and occupancy variations} \\
    \hline
    seat\_occupancy & initial\_seatbelt\_status & final\_seatbelt\_status & expected\_reminder\_status \\
    \hline
    Occupied & Unfastened & Fastened & FALSE \\
    \hline
    Occupied & Fastened & Unfastened & TRUE \\
    \hline
    Unoccupied & Unfastened & Fastened & FALSE \\
    \hline
    Unoccupied & Fastened & Unfastened & FALSE \\
    \hline
    \end{tabular}
\end{minipage}
}

\begin{figure}[h] 
    \centering
    \includegraphics[width=0.75\linewidth]{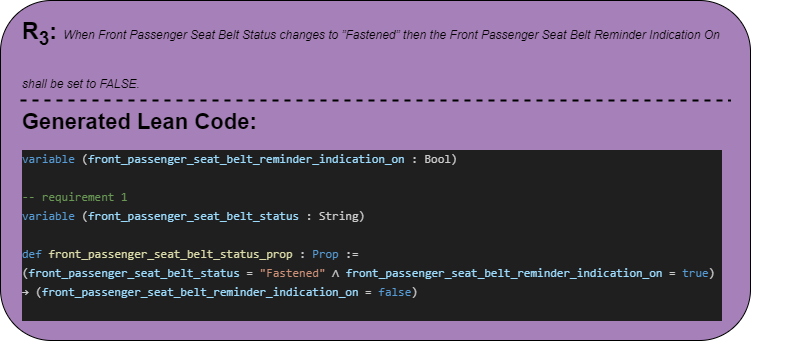} 
    \caption{Lean Formalization of Requirement $R_3$ }
    \label{fig:Sample R3 Output}
\end{figure}

\begin{figure}[h] 
    \centering
    \includegraphics[width=0.75\linewidth]{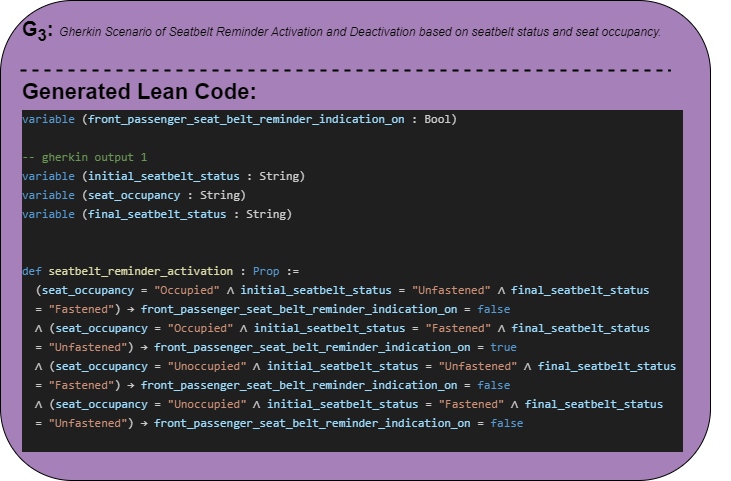} 
    \caption{Lean Formalization of Gherkin Scenario $G_3$ }
    \label{fig:sample G3 output}
\end{figure}

We hypothesize that the LLM-generated Gherkin scenario ($G_3$), due to its added complexity, will not be logically equivalent to the original requirement ($R_3$).We can see the output generated by DeepSeek-Prover on $R_3$ and $G_3$ and in figure \ref{fig:Sample R3 Output} and figure \ref{fig:sample G3 output} respectively.

In these set of examples, if we look closely we can observe that the LLM-generated Gherkin scenario $G_3$ contains a lot of extra variables that are not present in the original requirement $R_3$, hence they are logically inconsistent. Specifically speaking, the natural language requirement $R_3$ does not consist references of \textit{seat\_occupancy} and \textit{final\_seatbelt\_status}. So ideally, \textbf{the Lean code provided by the Autoformalizer LLM should either prove they are inconsistent or fail to run}.

\begin{figure}[h] 
    \centering
    \includegraphics[width=1\linewidth]{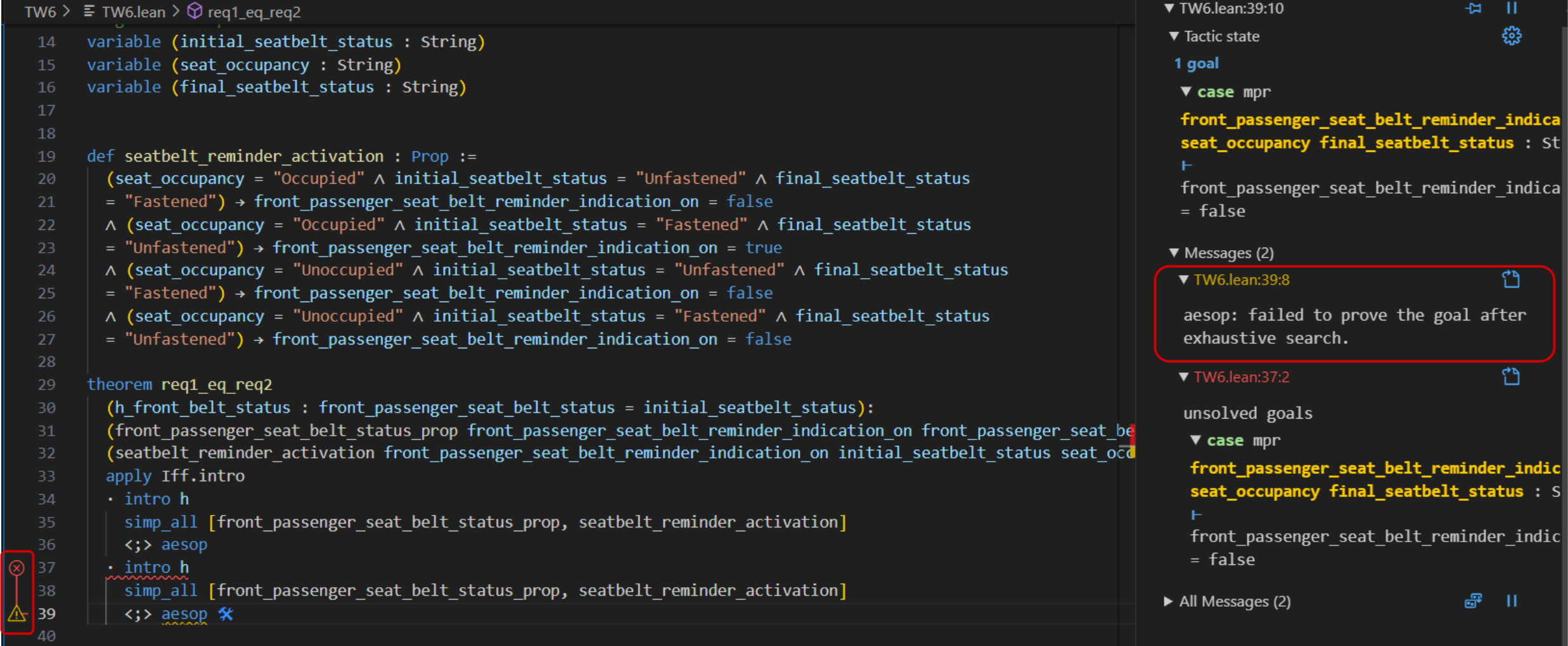} 
    \caption{Lean code fails to prove $R_3$ and $G_3$ are logically equivalent}
    \label{fig:r3-g3 logical inequivalence}
\end{figure}

The final result of the LLM-generated Lean 4 code in the Lean compiler is provided in Figure \ref{fig:r3-g3 logical inequivalence}.

\section{Discussion}
\label{sec:discussion}

The Autoformalization pipeline we explored in this paper to verify LLM-outputs is given by Figure \ref{fig:autoformalizer-pipeline}. In this pipeline, we do the following things:

\begin{figure}[ht] 
    \centering
    \includegraphics[width=1\linewidth]{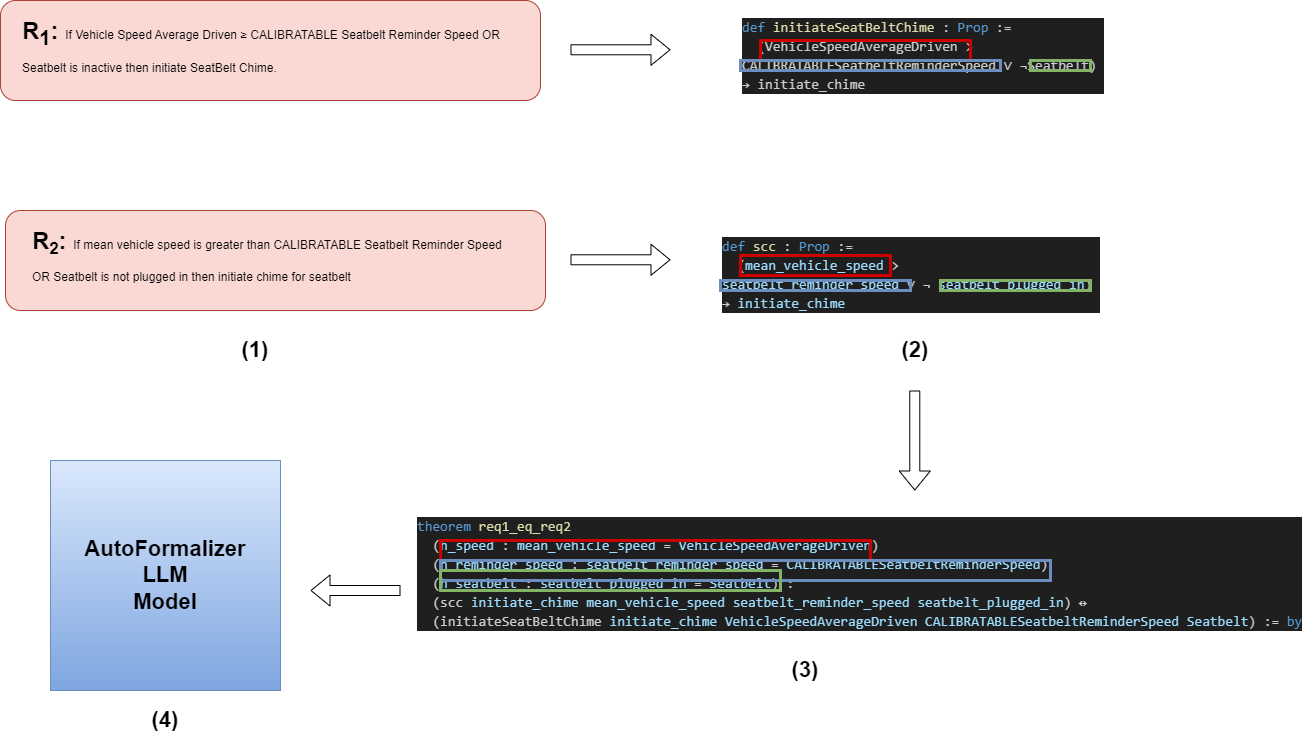} 
    \caption{Pipeline for Autoformalization. (1) are the NL Requirements. (2) is the proposition created in Lean by the LLM. (3) is how we manually ground the variables created by the propositions. (4) is to provide this manually grounded theorem to the Autoformalizer LLM to prove/disprove equivalence }
    \label{fig:autoformalizer-pipeline}
\end{figure}

\begin{enumerate}
    \item We begin with a natural language (NL) requirement and use an autoformalizer-based LLM to translate this informal statement into a formal mathematical proposition in Lean as shown by (1) and (2) in Fig \ref{fig:autoformalizer-pipeline}.
    \item We then use these generated propositions to ground the variables in a theorem, such that we define that a given set of generated variables logically point to the same hypothesis. This is the step we take to go from (2) to (3) in Fig \ref{fig:autoformalizer-pipeline}.
    \item The final, grounded theorem is then provided to the autoformalizer-based LLM. The model's purpose is to generate Lean code to prove the biconditional equivalence of the two propositions which is done in (4).
\end{enumerate}

The outcome of this step either yields a successful proof, thereby confirming logical equivalence, or a failed proof, which indicates that the statements are not logically consistent. The model's failure to complete a proof in Lean is a key indicator of a discrepancy, as it signals that the two statements cannot be reconciled under the given logical system.

\subsection{Formalization is dependent on the Language Model's interpretability}
Previous work\cite{weng2025autoformalization} has noted that informalizing a formal statement is often easier for LLMs than formalizing an informal statement. This difficulty stems from the autoregressive nature of current-generation LLMs, which are heavily reliant on token structure, making them well-suited for generation but less reliable for the precise, rigid syntax required by formal languages.

The core challenge lies in the mismatch between the processes: formalization is, by definition, non-ambiguous and rigid, yet it is being performed by an inherently non-deterministic, probabilistic model. While the flexibility of LLMs allows them to process the varied and informal nature of natural language inputs, the resulting formal output itself should not be subject to non-deterministic interpretation.This problem is magnified when the source material is ambiguous, a common characteristic of system requirements. The resulting formalized interpretation (Lean code, in our case) must accurately capture the precise intent of the requirement, including subtle system nuances.

For example, consider the requirement:

\textit{$R_4$: "When the seatbelt is Unfastened and the vehicle is in motion then the Front Passenger Seat Belt Reminder Indication On shall be set to TRUE." }

The condition "vehicle in motion" can have different interpretations depending on the system design (e.g., speed $ > $ 0, or gear not in park).

\begin{figure}[ht] 
    \centering
    \includegraphics[width=0.75\linewidth]{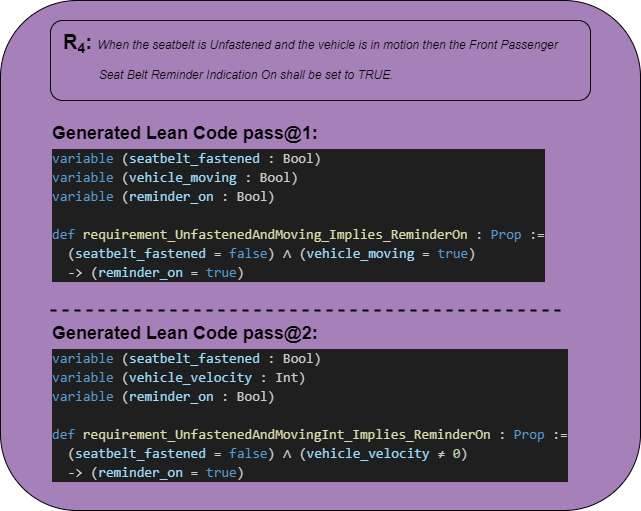} 
    \caption{Generated Lean Code for ambiguous informal statement }
    \label{fig:inconsistent-reqs}
\end{figure}

In Figure \ref{fig:inconsistent-reqs} theoretically both the generated lean code represent the input requirement properly given the amount of information about the system. However, they are not formally equivalent because they represent "vehicle in motion" differently: pass@1 represents the variable as a boolean whereas pass@2 represents the variable as an integer and hence it's not possible to formally say that both the code snippets are logically equivalent even though they were derived from the same informal requirement. This represents a flaw of a design solely relying on informal natural language requirements without being grounded in the system more deeply.

\subsection{Can Grounding of variables be Automated?}

In our autoformalization pipeline (Figure \ref{fig:autoformalizer-pipeline}, (2) $\rightarrow$ (3) $\rightarrow$ (4)), a critical and inescapable requirement is the grounding of overlapping variables to prove logical equivalence. This necessity arises because Large Language Models (LLMs) frequently interpret and name variables differently across separate formalizations, even when the underlying concept is the same. For instance, Figures \ref{fig:inconsistent-reqs}, \ref{fig:sample R1 output} and \ref{fig:sample R2 output} clearly demonstrate the issue, where a single physical parameter is represented using distinct variable names and types.

This challenge is a direct consequence of an LLM formalizer making assumptions based on limited contextual knowledge of the target system. The inconsistency highlights the necessity of grounding the entire autoformalization pipeline within the system's defined ontology to achieve reliable results.

Although variable grounding is a purely logical process and should ideally be done algorithmically to eliminate human error, it currently remains the most time-consuming step. We propose two primary directions for automating this semantic matching task:
\begin{enumerate}
    \item \textbf{Contextual Single-Pass Formalization:} A naive yet potentially effective approach is to autoformalize the given set of statements along with system information in a single, comprehensive pass. This helps ensure the LLM interprets and defines variables consistently. However, due to the LLM's non-deterministic nature, this cannot be guaranteed.
    \item \textbf{Similarity-Based Grounding:} A more robust approach is to explore similarity-based methods. The embeddings generated for the variables within the formal propositions can be computed and compared with the embeddings of established variables in the system's ontology (e.g., a formal data dictionary) to ensure accurate alignment.
\end{enumerate}

\subsection{Grounding as a verification mechanism}
There is significant scope for developing LLM-based solutions to automate grounding. However, automating a verification step using a probabilistic model introduces an ironic challenge: the need to develop a robust verification mechanism for the LLM's own variable-grounding hypothesis. This fundamental trade-off between efficiency and certainty is a central theme for future research in this domain.

Furthermore, the integration of grounding with autoformalization has broader applications for ensuring that LLM-generated Chain-of-Thought reasoning maintains logical consistency within constrained systems (e.g., smart-home systems, vehicle assistants). Since these systems have a limited set of defined actions, future work can focus on: developing a \textbf{knowledge graph} or similar data structure of branching system instructions, creating a mechanism for \textbf{LLMs to traverse this structure} while generating output \& \textbf{employing autoformalization} to guarantee that the LLM's generated output is logically consistent with the system's predefined capabilities.

\subsection{Future Research Questions}
\paragraph{RQ1:} \textbf{What is the most effective method for autoformalizing an informal statement?} Our findings suggest that a single-pass conversion may not always be accurate, especially for complex requirements. As shown in other studies \cite{li2024autoformalize}, generating multiple formalized statements and selecting the most probable one (k-pass) could lead to more reliable conversions.
\paragraph{RQ2:} \textbf{How can the crucial step of variable grounding and normalization be reliably automated given a set of informal statements?} Our experiment revealed that manual grounding of variables is the most critical and labor-intensive step in this pipeline. While some research suggests using rule-based techniques\cite{aavani2011grounding}, there is significant potential for LLMs to automate this step\cite{thompson2025grounding}. A key question is which approach is more reliable and scalable. 
\paragraph{RQ3:} \textbf{Can we use autoformalization to develop verification systems for LLMs in environments that require logical consistency?} In this paper, we have shown that the construction of such a formal verification pipeline is feasible, demonstrating its potential as a proof-of-concept for formally assessing the logical equivalence of LLM-generated outputs against original natural language requirements. However, we need to explore algorithms that formally verify autoregressive output against knowledge structures to ensure consistency.

\paragraph{}
Autoformalization shows significant potential for logically verifying concepts or ideas generated by probabilistic models. However, more research is needed to optimize this approach and address the challenges identified in this paper.

\section{Conclusion}
\label{sec:conclusion}

In this paper, we presented a preliminary exploration of two novel applications of autoformalization: the formal verification of LLM-generated outputs against natural language requirements and the checking of logical equivalence among requirements. Our study served as a \textbf{proof-of-concept}, outlining a pipeline that leverages an LLM-based autoformalizer to translate informal statements into formal Lean propositions and then utilizing the autoformalizer to formally prove the equivalence of generated propositions. Through two distinct experiments, we showed that this pipeline holds significant promise for improving the fidelity and consistency of content produced by probabilistic models.
While our findings are based on a limited sample, they suggest that autoformalization can play a pivotal role in bridging the gap between ambiguous natural language and rigorous, verifiable logic. We also identified several critical avenues for further investigation to generalize and strengthen such a pipeline.
Ultimately, this paper lays the groundwork for a formal verification framework. As autoformalization methods mature, we envision them becoming an integral practice for ensuring the correctness, reliability, and trustworthiness of LLM-powered applications.

\newpage
\begin{appendices}
\section{Prompt Templates}
    \label{app:prompt}
    \subsection{Prompt to Formalize Natural Language requirement or Scenario}
    \begin{lstlisting}[language=]
# Formalize the following requirement using Lean syntax:
{requirement}

Translate the natural language requirement into a formal Lean 4 proposition. The result should be a `def` statement that defines a proposition (`Prop`).

Complete the following code and substitute the brackets with appropriate variables from requirements:

The definition should have the following structure:

```lean4
-- define variables here
variable (<VARIABLE> : <TYPE>)
variable (<VARIABLE> : <TYPE>)

def <ACTION_name_FOR_function> : Prop :=
-- include all conditions in the given here and finally they should either imply or not imply the action
(CONDITION A \land CONDITION B \land ...) -> ACTION
        ```
    \end{lstlisting}
    \subsection{Prompt to Prove Theorems}
    \begin{lstlisting}[language=]
Given the following Lean code, reason out and finally prove that either:
- the given two functions are logically equivalent and consistent
- the given two functions are inconsistent

Continue and complete the theorem based on the provided code:

```lean4
import Mathlib.Data.Real.Basic

variable (front_passenger_seat_belt_reminder_indication_on : Bool)

-- requirement 1
{requirement_proposition}

-- gherkin output 1
{gherkin_output_proposition}

theorem req1_eq_req2
(h_front_belt_status : front_passenger_seat_belt_status = initial_seatbelt_status):
(front_passenger_seat_belt_status_prop front_passenger_seat_belt_reminder_indication_on front_passenger_seat_belt_status) <->
(seatbelt_reminder_activation front_passenger_seat_belt_reminder_indication_on initial_seatbelt_status seat_occupancy final_seatbelt_status) := by
    \end{lstlisting}
\end{appendices}
\newpage
\bibliographystyle{unsrtnat}
\bibliography{template}
\end{document}